\pdfoutput=1

\documentclass[11pt]{article}

\usepackage[final]{acl}

\usepackage{times}
\usepackage{latexsym}
\usepackage{hyperref}

\usepackage[T1]{fontenc}

\usepackage[utf8]{inputenc}

\usepackage{microtype}

\usepackage{inconsolata}

\usepackage{graphicx}

\usepackage{url}
\usepackage{multirow} 
\usepackage{algorithm}
\usepackage{algpseudocode}
\usepackage{amsmath}
\usepackage{amssymb}
\DeclareMathOperator*{\argmax}{arg\,max}
\title{EMORL: Ensemble Multi-Objective Reinforcement Learning for Efficient and Flexible LLM Fine-Tuning}

\author{
 \textbf{Lingxiao Kong\textsuperscript{1,*}},
 \textbf{Cong Yang\textsuperscript{2}},
 \textbf{Susanne Neufang\textsuperscript{3}}, \\
 \textbf{Oya Deniz Beyan\textsuperscript{1,3}},
 \textbf{Zeyd Boukhers\textsuperscript{1,3,*}}
\\
 \textsuperscript{1}Fraunhofer Institute for Applied Information Technology FIT, Germany \\
 \textsuperscript{2}Soochow University, China \\
 \textsuperscript{3}University Hospital of Cologne, Germany
\\
\texttt{\{\href{mailto:lingxiao.kong@fit.fraunhofer.de}{lingxiao.kong}, \href{mailto:oya.deniz.beyan@fit.fraunhofer.de}{oya.deniz.beyan}, \href{mailto:zeyd.boukhers@fit.fraunhofer.de}{zeyd.boukhers}\}@fit.fraunhofer.de}\\
\texttt{\href{mailto:susanne.neufang@uk-koeln.de}{susanne.neufang@uk-koeln.de}}\\
\texttt{\href{mailto:cong.yang@suda.edu.cn}{cong.yang@suda.edu.cn}}
\\
\\
}

\begin{document}
\maketitle
\begin{abstract}

Recent advances in reinforcement learning (RL) for large language model (LLM) fine-tuning show promise in addressing multi-objective tasks but still face significant challenges, including competing objective balancing, low training efficiency, poor scalability, and limited explainability. Leveraging ensemble learning principles, we introduce an Ensemble Multi-Objective RL (EMORL) framework that fine-tunes multiple models with individual objectives while optimizing their aggregation after the fine-tuning to improve efficiency and flexibility. Our method is the first to aggregate the hidden states of individual models, incorporating contextual information from multiple objectives. This approach is supported by a hierarchical grid search algorithm that identifies optimal weighted combinations. We evaluate EMORL on counselor reflection generation tasks, using text classification models to score the generations and provide rewards during RL fine-tuning. Through comprehensive experiments on the PAIR and Psych8k datasets, we demonstrate the advantages of EMORL against existing baselines: significantly lower and more stable training consumption ($17,529\pm 1,650$ data points and $6,573\pm 147.43$ seconds), improved scalability and explainability, and comparable performance across multiple objectives.

\end{abstract}

\begingroup
\renewcommand\thefootnote{}\footnotetext{\textsuperscript{*} Corresponding authors}

\endgroup

\section{Introduction}

\begin{figure}[t!]
   \begin{center}
\includegraphics[width=1.0\linewidth]{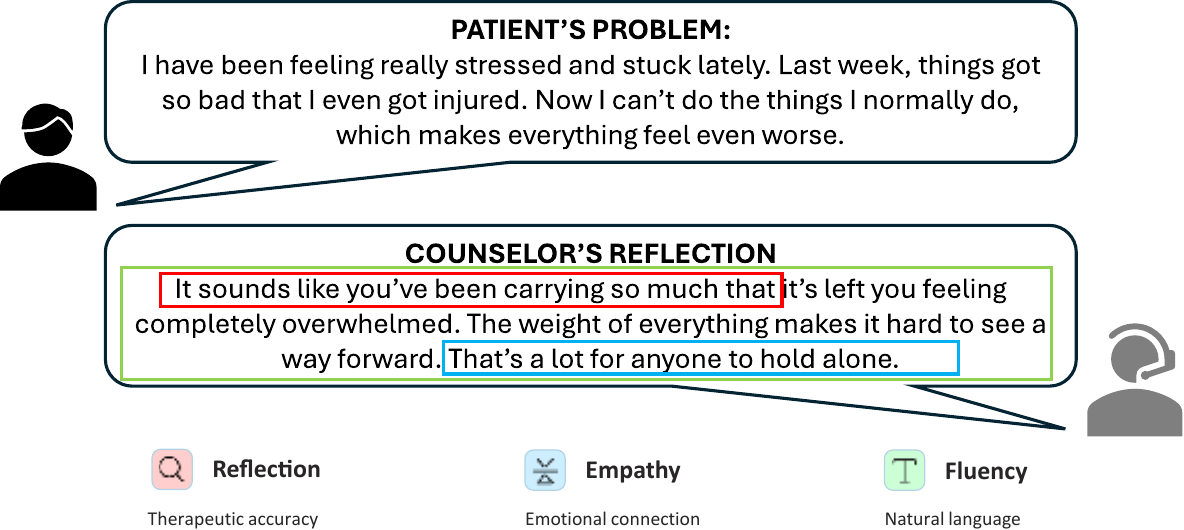}
   \end{center}
   \caption{The counselor reflection generation task requires creating reflective, empathetic, and fluent statements in response to the patient's problem. }
   \label{fig: task}
\end{figure}

Multi-objective optimization (MOO) for large language models (LLMs) is a crucial research direction in natural language processing (NLP) tasks that require fulfilling diverse and often competing requirements~\citep{vaswani2017attention, qin2024large, kumar2024large}. Such tasks require models to simultaneously optimize for multiple evaluation criteria, balancing trade-offs between different objectives. This paper focuses on the counselor reflection generation task, where LLM reflects on users' prompts, building the first step of therapeutic communication~\citep{o2023automatic}. As exemplified in Figure~\ref{fig: task}, this task requires generations that optimize competing objectives, such as \emph{reflection}, \emph{empathy}, and \emph{fluency} in high-quality counseling interactions. 

Reinforcement learning (RL) offers a promising approach for fine-tuning pre-trained LLMs in MOO~\citep{lin2024reinforcement, parthasarathy2024ultimate}. Conventional RL fine-tuning approaches combine multiple objectives into one reward function, enabling optimization of objectives during training~\citep{okano2023generating, perez2024dynamic, dann2023reinforcement}. However, these approaches face significant challenges: optimizing appropriate weights for each objective, maintaining training efficiency, and ensuring scalability while preserving result explainability~\citep{hayes2022practical, dulac2021challenges}. From another perspective, compared to RL with Human Feedback (RLHF)~\citep{ouyang2022training}, MOO explicitly defines objectives, advantaging user-specified LLM characteristics. Human feedback typically expresses preferences rather than specific objective values, requiring exhaustive human annotations, while MOO directly targets explicit objectives, transcending implicit instruction limitations. Challenges also exist in ensuring objectives align with human perception and in incorporating many objectives simultaneously. These challenges highlight the need for more efficient and flexible fine-tuning methods.

\begin{figure}[t!]
   \begin{center}
\includegraphics[width=1.0\linewidth]{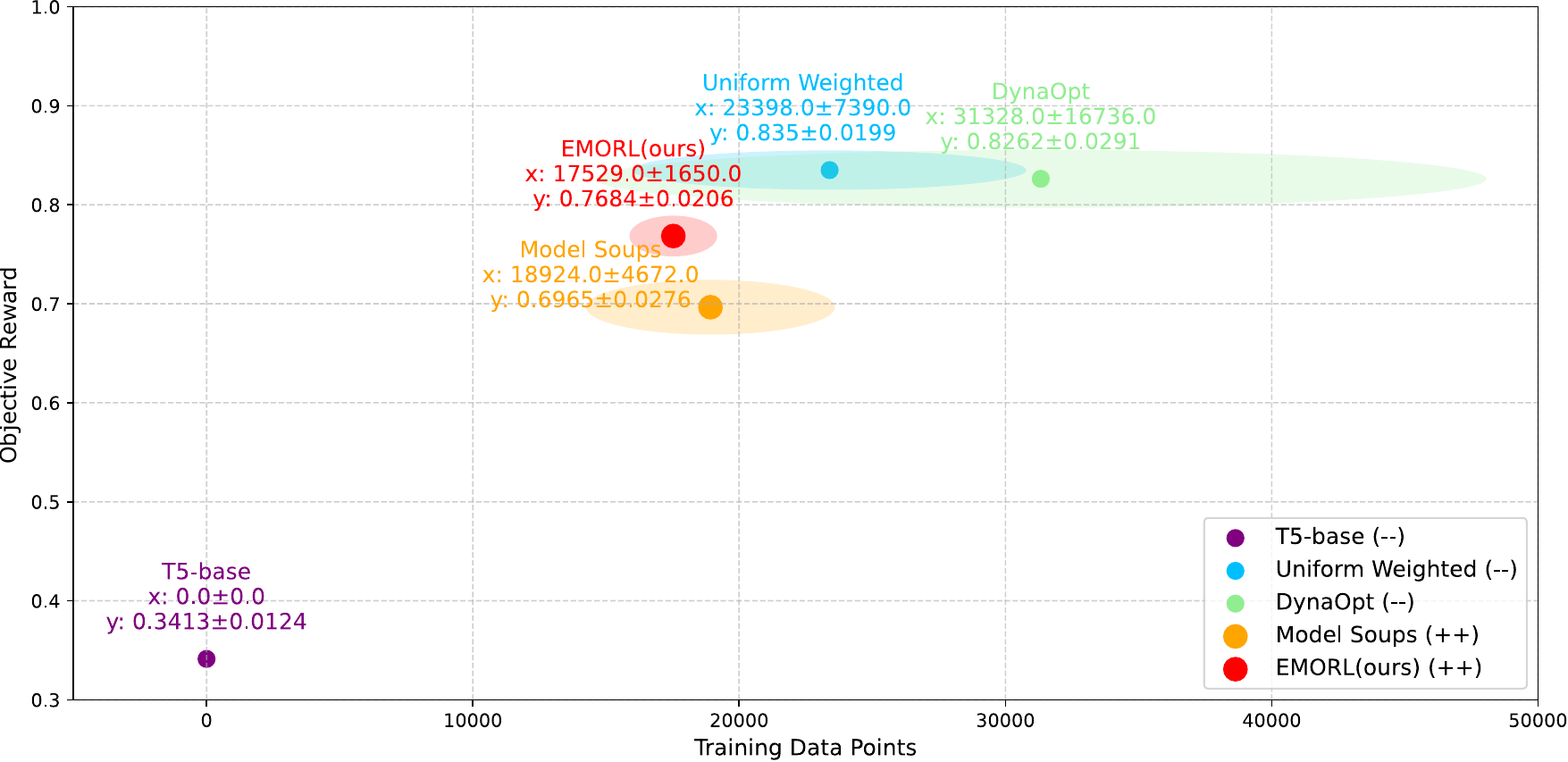}
   \end{center}
   \caption{Comparison of training consumption (x-axis) and objective reward (y-axis). Point size and "+/-" indicate scalability and explainability capabilities. EMORL achieves comparable performance with higher efficiency while maintaining better scalability and explainability.}
   \label{fig: scatter}
\end{figure}

Ensemble learning offers a potential solution for this growing need by aggregating multiple trained models into a global model~\citep{vanhaesebrouck2017decentralized}. We propose a novel Ensemble Multi-Objective RL (\textbf{EMORL}) framework for LLM fine-tuning that distributes objectives to respective models, enabling training for individual objectives independently. Our experiments demonstrate that models trained with single-objective reward functions converge significantly faster than those optimizing multiple objectives simultaneously. The framework incorporates a hidden states aggregation method coupled with a hierarchical grid search algorithm during aggregation, which efficiently identifies optimal weights for combining these single-objective models. Our results demonstrate that the aggregated output achieves higher training efficiency while achieving performance comparable to models trained using single-policy methods, as shown in Figure~\ref{fig: scatter}. The performance is represented by an objective reward, which uses a utility function that weights objectives according to user-specific preferences. Our framework is highly scalable, accommodating different user preferences and new objectives through aggregation. Additionally, the framework enhances explainability by providing insights into the importance of different objectives easily through evaluating different weighted combinations of objectives on test samples. The code for our framework and experiments is publicly available and can be found at \url{https://github.com/engineerkong/EMORL}.

Succinctly, our main contributions are as follows: (1) We introduce \textbf{EMORL} framework that separately trains and aggregates models in the counselor reflection generation task. (2) We develop an effective hidden-state level aggregation method and a hierarchical grid search algorithm for optimizing the weighted combination. (3) We demonstrate our framework's effectiveness through a comprehensive evaluation against MOO baselines, achieving comparable performance across multiple objectives while offering lower and more stable training resources, improved scalability and explainability.

\section{Related Work}

Prior work on RL for LLM fine-tuning has explored various approaches to balance multiple objectives. The most common methods include single-policy approaches that learn all objectives simultaneously using scalarized reward functions, and multi-policy approaches that train and populate numerous policies with vectorized rewards to achieve Pareto optimality. To address the efficiency and flexibility limitations of these existing methods, we analyze related single-policy and multi-policy techniques and present meta-policy methods with ensemble learning, which combines multiple policies to create a meta-policy that better addresses MOO.

\subsection{Single-Policy}
With prior knowledge, objectives are combined into one reward function or loss function using fixed weights $\lambda_i$ for individual objectives, as shown in Equation~\eqref{eq: single-policy}, where the total reward $R_{total}$ is summed up using respective rewards $R_i$.

\begin{equation}
    R_{total} = \lambda_1 \cdot R_1 + ... + \lambda_n \cdot R_n
    \label{eq: single-policy}
\end{equation}

\citet{ziegler2019fine} explored RL for LLM fine-tuning using fixed weights to combine task-specific rewards with auxiliary objectives like fluency. However, selecting appropriate weights to balance objectives effectively is challenging, since it requires extensive trial-and-error to verify the effectiveness of weighted combinations. \citet{mohan2023autorl} proposed AutoRL to automate the selection of optimal weights. However, it still requires training numerous candidate policies to trace back the optima, remaining computationally intensive. Dynamic weighting approaches like \citet{perez2024dynamic}'s Multi-Armed Bandit and \citet{pu2024dynamic}'s Markov Chain strategies enable continuous adaptation based on received rewards. These methods adjust objective weights during training based on the model's performance, context, or external feedback, enabling a more flexible and adaptive balance between competing goals. However, this approach requires these additional mechanisms to adjust weights, which increases computational complexity and introduces training instability. 


\subsection{Multi-Policy}
Multi-policy methods train multiple models simultaneously and generate Pareto-optimal solution sets using evolutionary algorithms or other population-based mechanisms~\citep{vamplew2022scalar}. These methods utilize vectorized rewards as shown in Equation~\eqref{eq: multi-policy} rather than scalarized reward functions. Since vectorized rewards encode no user preferences, the methods first iteratively obtain diverse Pareto-optimal solutions, then apply a user preference to select a single optimal solution.

\begin{equation}
    \mathbf{R_{total}} = \begin{bmatrix} R_1 & R_2 & \cdots & R_n \end{bmatrix}^\top
    \label{eq: multi-policy}
\end{equation}

These approaches offer a more comprehensive exploration of the solution space compared to single-policy methods. However, they require training numerous models and computationally expensive population mechanisms, making them particularly challenging for large-scale models like LLMs~\citep{hayes2022practical}. 

\subsection{Meta-Policy}

To address these inflexibilities and inefficiencies, meta-policy approaches use ensemble learning to enhance model adaptability and reduce computational overhead~\citep{yuan2025towards}. Ensemble learning comprises three main approaches: bagging (combining separately trained models), boosting (sequential training to improve upon previous models), and stacking (using a meta-learner to integrate outputs from diverse models, requiring additional meta-training)~\citep{song2023ensemble}. Meta-policy approaches divide MOO into two hierarchical phases~\citep{zhang2022meta}. At the \textbf{lower level}, several models are trained according to different sampled user preferences. At the \textbf{higher level}, these models are combined to address specific user preferences and produce a single optimal solution. These approaches shift objective balancing from the lower level to the higher level, making multi-objective optimization more flexible and efficient.

\citet{wortsman2022model} demonstrated Model Soups that combines the parameters from multiple trained models using bagging. Their learned soup approach learns the weights for parameter-level aggregation and aggregates multiple models into a single model through gradient-based optimization. \citet{shi2024decoding} combined the predicted tokens from models trained on individual objectives, namely logit-level aggregation, and leveraged Legendre transformation and f-divergence to optimize the performance of combined predictions for NLP tasks. \citet{feriani2022multiobjective} balanced the minimum throughput and standard deviation in load balancing by learning a meta-policy using distilled data from multiple trained models. \citet{malik2024attention} trained a meta-policy by stacking three types of feature embedding representations from different trained models and used a multi-head attention mechanism to non-linearly combine the features for scoring sentiment analysis.

However, as demonstrated in Appendix~\ref{subsec: params agg} and ~\ref{subsec: logits agg}, parameter-level aggregation and logit-level aggregation underperform in MOO, particularly when the objectives differ significantly. This indicates that the application of ensemble learning in this multi-objective context remains underexplored. In light of this, we explore the novel hidden-state level aggregation by combining the contextual information from individual objectives to achieve efficient and flexible multi-objective LLM fine-tuning, while maintaining comparable performance.

\section{Challenges}

In this section, we analyze the challenges of MOO by comparing the training processes of models with individual objectives against multi-objective models using conventional single-policy methods. We tested five fine-tuning setups for counselor reflection generation to demonstrate these challenges, focusing on the objectives: \emph{reflection}, \emph{empathy}, and \emph{fluency}. Three setups optimized single objectives separately, while two single-policy approaches fine-tuned for all three objectives: (1) Uniform Weighted, assigning equal weights ($\frac{1}{3}$) to each objective, and (2) DynaOpt, dynamically adjusting weights using a multi-armed bandit algorithm. Experiments used $5$ random seeds with $3$ generation runs each, evaluating objective rewards, data consumption, and training time. Progress was tracked via Weights \& Biases, plotting objective reward against data consumption.

\begin{figure}[t]
   \begin{center}
   \includegraphics[width=1.0\linewidth]{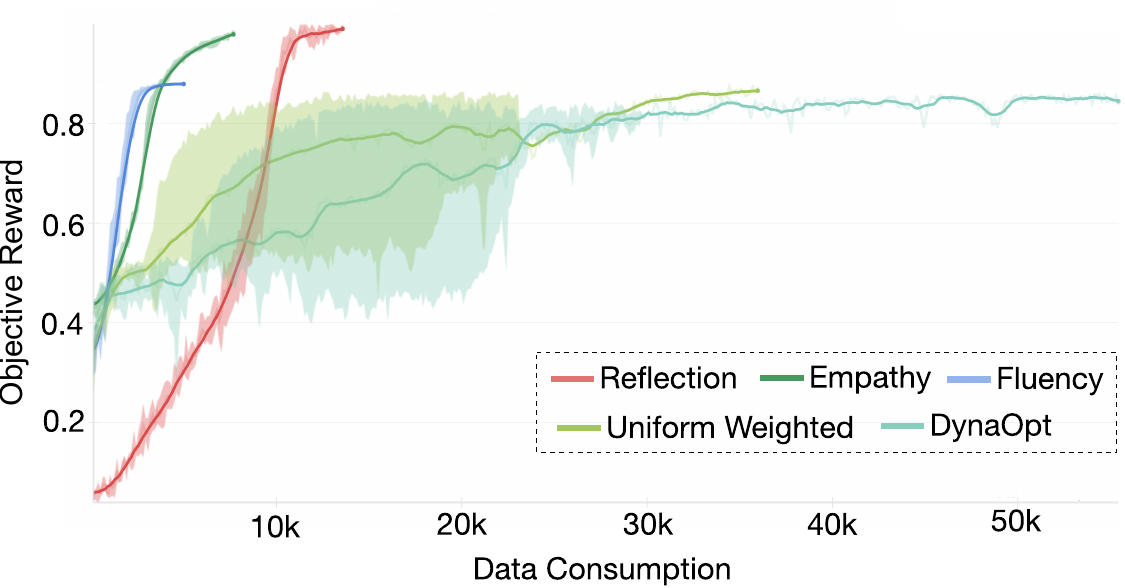}
   \end{center}
   \caption{RL fine-tuning processes logs for $5$ setups, highlighting single-objective models' advantages in convergence speed, process stability, and performance.}
   \label{fig: fine-tuning wandb}
\end{figure}

As shown in Figure~\ref{fig: fine-tuning wandb} and the detailed results in Table~\ref{tab: fine-tuning detail} (Section~\ref{sec: appendix}), our results highlight key differences between single-objective and conventional MOO in three aspects. First, \textbf{convergence speed}: single-objective models converged faster, with fluency models achieving the quickest convergence ($4,809$ data points, $1,629.19$ seconds). Multi-objective models were significantly slower, with Uniform Weighted requiring $23,398$ data points and $5,967.84$ seconds, and DynaOpt needing $31,328$ data points and $8,029.15$ seconds due to the overhead of dynamic weighting. Second, \textbf{process stability}: single-objective fine-tuning showed consistent convergence with minimal variation ($\pm 335$ data points, $\pm 104.40$ seconds for reflection). Multi-objective models exhibited less stability, with Uniform Weighted varying by $\pm 7,390$ data points and $\pm 1,875.50$ seconds, and DynaOpt by $\pm 16,736$ data points and $\pm 4,365.64$ seconds, reflecting the complexity of balancing multiple objectives during training. Third, \textbf{performance metrics}: single-objective models achieved higher objective rewards (approaching $1.0$ for reflection and empathy), while multi-objective models averaged below $0.85$, indicating inherent performance trade-offs in optimizing multiple objectives.

These observations suggest that integrating multiple objectives in training inherently presents convergence, stability, and performance challenges. A promising research direction emerges from this insight: \textbf{ensembling single-objective models to achieve efficient and flexible MOO}.

\section{Methodology}

\begin{figure*}[t!]
\begin{center}
\includegraphics[width=0.95\linewidth]{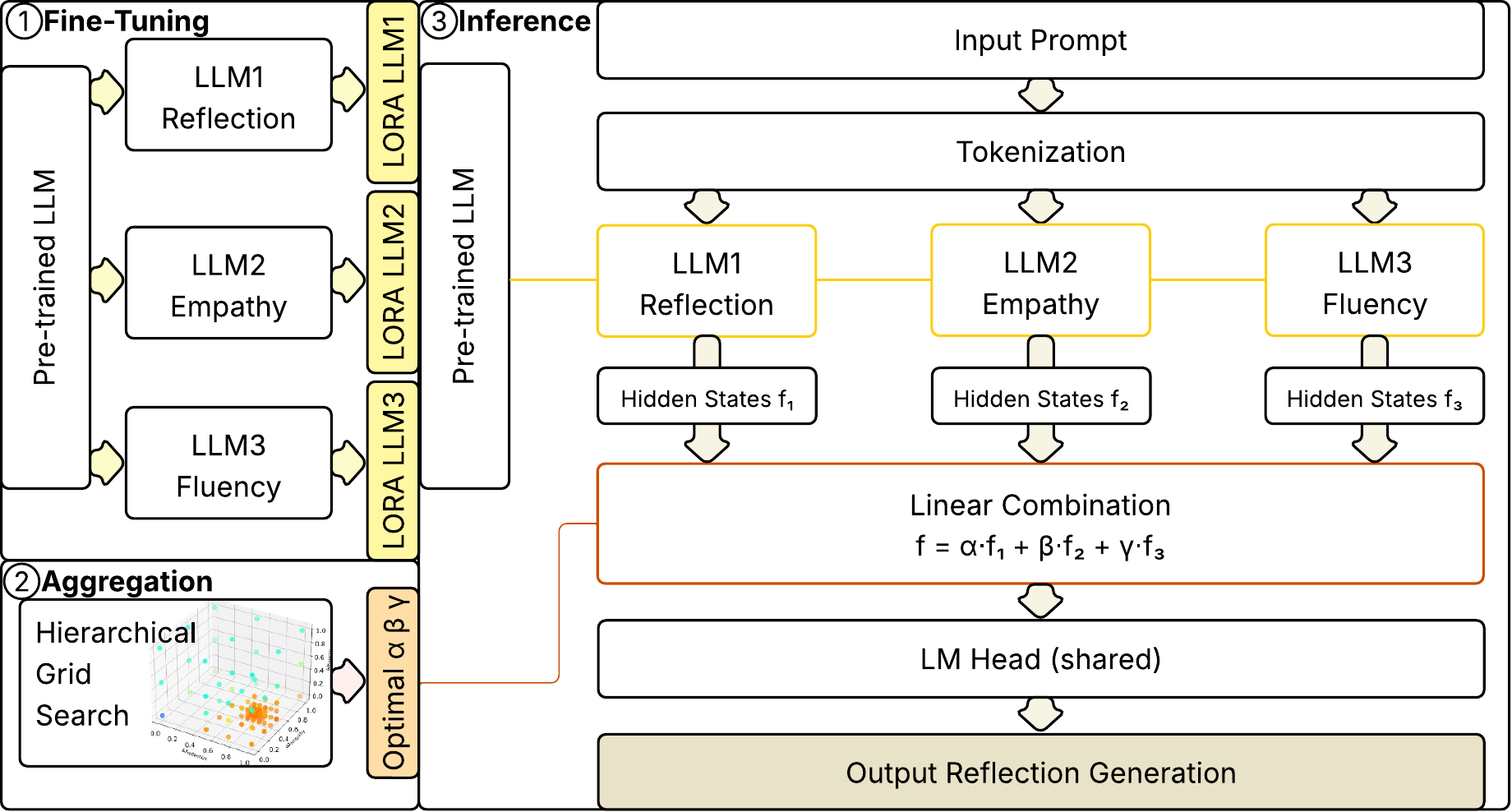}
\end{center}
\caption{The EMORL framework illustrates a three-stage process: fine-tuning, aggregation and inference. }
\label{fig: EMORL}
\end{figure*}

As shown in Figure~\ref{fig: EMORL}, the EMORL framework consists of three key stages. First, multiple models are independently fine-tuned for distinct individual objectives. Second, we employ hierarchical grid search to find optimal linear combinations according to our utility function, which defaults to average objective preferences. Finally, we use the optimal weights to aggregate models at the hidden-state level during inference, generating outputs that effectively integrate all objectives. This ensemble learning approach simplifies the complex multi-objective fine-tuning problem into an optimization problem for aggregation weights.

\subsection{Hidden-State Level Aggregation}
The decoder of an LLM generates hidden states that capture high-level features for contextual understanding, semantic representations, cross-attention patterns, and task-specific information~\citep{raffel2020exploring}. Typically, the last hidden states are processed by a language model head to compute logits, representing token probabilities across the vocabulary list. Logits are then used to generate tokens via the $\argmax$ operation. In our approach, we aggregate the last hidden states from multiple objective-specific models to cohesively integrate their high-level features using a linear combination, as formulated in Equation~\eqref{eq: aggregation}. 

\begin{equation}
    \begin{gathered}
    \mathbf{f} = \alpha \cdot \mathbf{f}_1 + \beta \cdot \mathbf{f}_2 + \gamma \cdot \mathbf{f}_3, \text{where} \; \alpha, \beta, \gamma \in [0, 1] \\
    \mathbf{h}_t = H_{LM}(\mathbf{f}) \in \mathbb{R}^{|V|} \\
    \text{token}_t = \underset{i \in \{1, 2, \ldots, |V|\}}{\arg\max} \; \mathbf{h}_t[i]
    \end{gathered}
    \label{eq: aggregation}
\end{equation}

Where $\mathbf{f}_i$ represents the hidden state from the $i$-th model, $\mathbf{f}$ is the linearly aggregated hidden representation, $H_{LM}$ denotes the language model head transformation that projects the hidden state to the vocabulary space, $\mathbf{h}_t$ is the resulting logits vector of dimension $|V|$ (vocabulary size), and $\text{token}_t$ is the selected token with maximum probability at timestep $t$. The coefficients $\alpha$, $\beta$, and $\gamma$ are weights constrained to the range $[0, 1]$, representing the independent contribution strength of each model.

In contrast to parameter-level and logit-level aggregation methods, whose experimental results we present in Appendix~\ref{subsec: params agg} and~\ref{subsec: logits agg} respectively, our hidden-state level aggregation approach ensures more consistent text generation while effectively incorporating features from all objective-specific models. The weight coefficients in this linear combination determine each objective's contribution of contextual information to the final output.

\subsection{Hierarchical Grid Search}
To effectively and efficiently search for the optimal weights $\alpha$, $\beta$, and $\gamma$, we experimented with various optimization methods. The principled Bayesian optimization approach described in the Appendix~\ref{subsec: bayesian optimization} exhibited slow convergence due to sampling and populating numerous evaluations, often trapped in local optima. Through analyzing performance distributions, we observed that the optima consistently appeared within higher-performance regions, leading us to explore the grid search methods.

We propose a hierarchical grid search algorithm that incorporates binary search concepts to reduce the computational complexity inherent in standard grid search~\citep{bishop2006pattern, gautschi2011numerical}. The developed hierarchical grid search achieves a computational complexity of $O(3^d \cdot \log_{2}\frac{1}{N})$ compared to grid search's $O(\frac{1}{N}^d)$, where $d$ represents the objectives and $N$ the precision level. The complexity comparison among these methods is shown in Figure~\ref{fig: complexity}. With three objectives and a precision level of $0.03125$, the required number of evaluations is reduced to $135$, compared to $32,768$ in standard grid search.

As detailed in Algorithm~\ref{algo: hierarchical search}, we first define the utility function addressing our preferences for objectives. By default, we set the utility to average the three objective scores. Our hierarchical approach divides the search axes for individual objectives into $3$ parts, creating $3^d$ initial grid points. We evaluate generation performance at these points and identify the most promising region by finding the $2^d$ cube with the highest total performance that maximizes our utility function. This region becomes the next search space, and we iterate these processes of grid generation and space refinement. The algorithm progressively focuses on smaller, more promising regions, proving particularly effective for this aggregation case.

\begin{algorithm}
\small
\caption{Hierarchical Grid Search}
\begin{algorithmic}[1]
\Require{utility function $f$, number of components $N$, iterations $I$, initial bounds $B_0 = [(0,1)]^N$}
\Ensure{Best point $p^*$, Best score $s^*$}

\State $p^* \gets \text{null}$
\State $s^* \gets -\infty$
\State $B_\text{current} \gets B_0$

\For{$\text{iter} = 1$ \textbf{to} $I$}
    \State $\text{grid\_points} \gets \text{GenerateGrid}(B_\text{current})$
    \State $\text{results} \gets \{\}$
    \ForAll{point $p \in \text{grid\_points}$}
        \State $\text{results}[p] \gets f(p)$
    \EndFor
    
    \State $p_\text{current} \gets \argmax(\text{results})$
    \If{$\text{results}[p_\text{current}] > s^*$}
        \State $s^* \gets \text{results}[p_\text{current}]$
        \State $p^* \gets p_\text{current}$
    \EndIf
    
    \State $\text{region} \gets \text{FindBestRegion}(\text{results})$
    \State $B_\text{current} \gets \text{ComputeBounds}(\text{region})$
\EndFor
\State \Return $p^*, s^*$
\end{algorithmic}
\label{algo: hierarchical search}
\end{algorithm}

\section{Experiments}

\subsection{Models}

To evaluate EMORL, we employ \textbf{T5-base}\footnote{https://huggingface.co/google-t5/t5-base} (220M parameters, Encoder-Decoder architecture) as the pre-trained model. \citet{perez2024dynamic} demonstrated T5's effectiveness for understanding and generating text in counseling tasks, making it suitable for our experiments compared to existing baselines. We utilize Self-Critical Sequence Training (\textbf{SCST}) as the RL algorithm, which generates candidate outputs and computes their mean reward as a baseline, thereby encouraging outputs to exceed this baseline performance~\citep{laban2021keep}. KL-divergence is incorporated in the loss function to constrain the fine-tuned model from diverging too far from the reference pre-trained model.

In the experiments, we fine-tune models using the default average utility function for \emph{reflection}, \emph{empathy}, and \emph{fluency} across $5$ random seeds, with $3$ generation runs, a training batch size of $16$, and up to $10,000$ steps while implementing early stopping at model convergence. We also employ LoRA~\cite{hu2021lora} to efficiently manage parameter updates by representing them via low-rank matrices and a scaling factor, as illustrated in Equation~\eqref{eq: lora_update}. Thus, we load the pre-trained model only once during inference and apply the LoRA parameters to update the pre-trained model toward each objective, significantly reducing the computational burden.

For evaluation, we compared EMORL models against four baselines: T5-base, Uniform Weighted, DynaOpt~\citep{perez2024dynamic}, and Model Soups~\citep{wortsman2022model}. All experiments were conducted on a Tesla V100 GPU with 32GB memory, 8 CPU cores, and 40GB system memory, with detailed consumption reported in Table~\ref{tab: evaluation metrics}.

\subsection{Datasets}
\textbf{Counselor reflection generation} task is a single-turn task to generate therapist-like reflective responses that accurately paraphrase and validate a client's expressed thoughts and feelings~\citep{o2023automatic}. The responses are expected to be reflective, empathic to the problem, and maintain fluency, as depicted in Figure~\ref{fig: task}. These generation objectives frequently conflict with one another. For instance, prioritizing accurate reflection of patient concerns makes maintaining empathetic tone challenging when addressing distressing truths. Similarly, excessive empathetic language may compromise coherence with other response components, undermining overall fluency and discourse quality.


The \textbf{PAIR}\footnote{https://lit.eecs.umich.edu/downloads.html} dataset is our primary dataset, split into a ratio of $80\%$, $10\%$, and $10\%$, for fine-tuning, aggregation, and inference, respectively~\citep{min-etal-2022-pair}. It contains $2,544$ single-turn client-counselor exchanges, covering topics ranging from mental health to lifestyle concerns like diet, exercise, and personal development. To assess the models robustly, we also conduct evaluations on the \textbf{Psych8k}\footnote{https://huggingface.co/datasets/EmoCareAI/Psych8k} dataset, sampling $10\%$ of its $8,187$ conversation pairs for inference. This dataset focuses on mental health interactions, including anxiety, depression, relationship issues, and stress management. It is widely used for training and evaluating LLMs in mental health counseling, and we leverage it here for reflection generation. The statistics of the datasets are presented in Table~\ref{tab: datasets statistics} (Section~\ref{sec: appendix}).


\begin{figure*}[t]
\begin{center}
\includegraphics[width=1.0\linewidth]{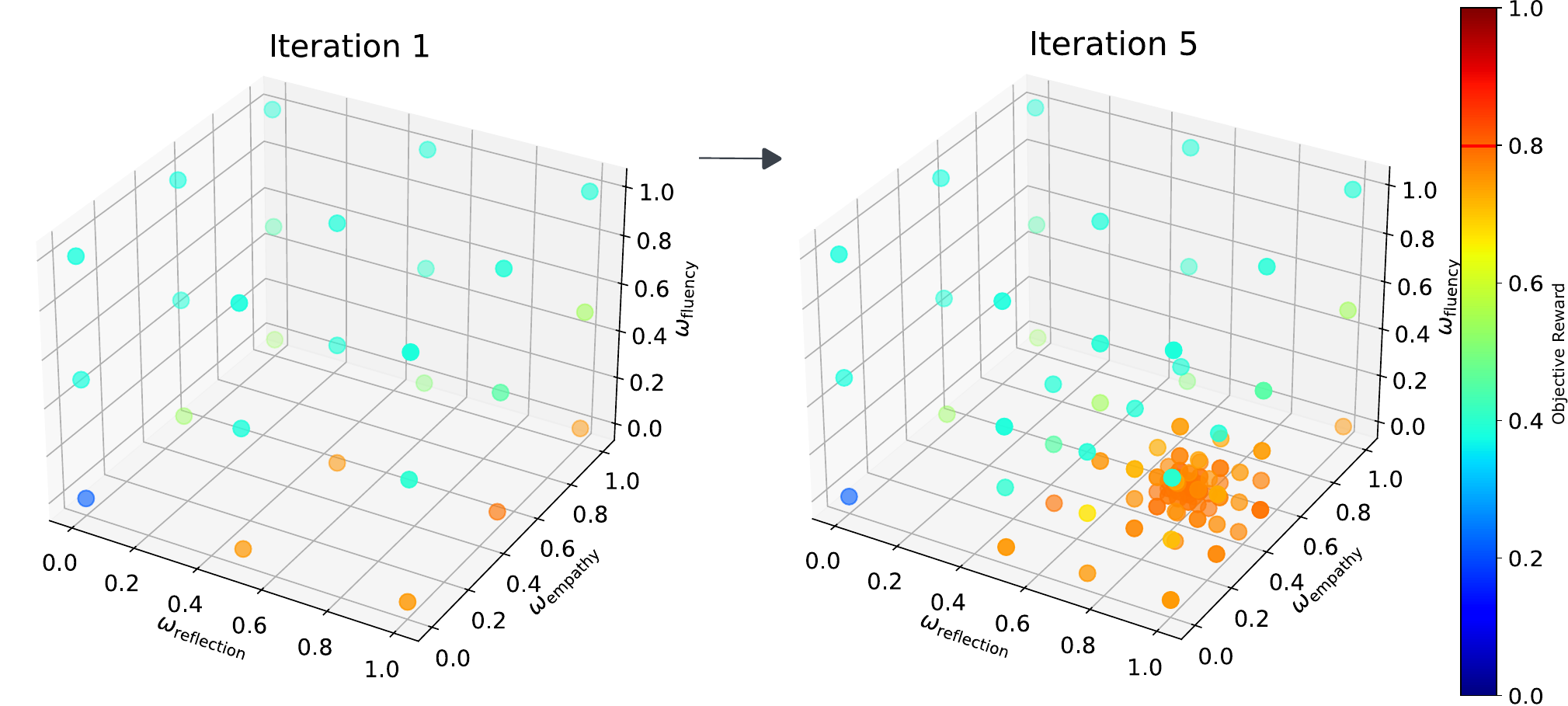}
\end{center}
\caption{The visualization illustrates the hierarchical grid search process, showing the transition from broad search spaces to more refined spaces, where optimal weight combinations are identified. The red line on the color map indicates the maximum objective reward achieved during the search. }
\label{fig: visualization}
\end{figure*}

\subsection{Metrics}

We evaluate comprehensive metrics beyond performance, focusing on six key aspects: (1) \textbf{Diversity-2} measures linguistic diversity; (2) \textbf{Edit rate} quantifies the avoidance of verbatim repetition; (3) \textbf{Data consumption} tracks the cumulative number of training samples processed; (4) \textbf{Time consumption} records the wall-clock time for each training iteration; (5) \textbf{Scalability} assesses the model's ability and flexibility to incorporate new preferences and additional objectives; (6) \textbf{Explainability} examines the interpretability of how each objective contributes to the final output.

For performance metrics, we employ text classification LLMs to score objectives on a scale from $0.0$ to $1.0$: (1) \textbf{Reflection} is assessed by the "roberta-base\footnote{https://huggingface.co/FacebookAI/roberta-base}" with checkpoints from \cite{perez2024dynamic}, which evaluates the relevance and contextual appropriateness. (2) \textbf{Empathy} is measured by the "bert-empathy\footnote{https://huggingface.co/MoaazZaki/bert-empathy}", which gauges emotional resonance and understanding. (3) \textbf{Fluency} is evaluated using "gpt2\footnote{https://huggingface.co/openai-community/gpt2}" by computing the inverse of perplexity, ensuring linguistic smoothness.

We conducted human evaluation of $640$ generations sampled from five models (T5-base, Uniform Weighted, DynaOpt, Model Soups and EMORL) across two datasets (PAIR and Psych8k). Two mental health experts independently rated each response on three performance metrics using a 3-point scale, normalized to 0.0-1.0: (1) \textbf{Reflection}: 0 (no reflection), 1 (simple mirroring), 2 (complex interpretation); (2) \textbf{Empathy}: 0 (no emotional awareness), 1 (basic understanding), 2 (deep emotional resonance); (3) \textbf{Fluency}: 0 (poor coherence), 1 (clear but awkward), 2 (natural and clear). The assessment instruction is detailed in Appendix~\ref{subsec: instruction}. 


\section{Results and Discussion}

\begin{table*}[t]
\centering
\small
\caption{The comprehensive metrics highlight measures beyond performance. The results demonstrate that our EMORL framework offers advantages in generation diversity, low training consumption, enhanced scalability, and improved explainability, outperforming other methods. }
\begin{tabular}{llllll}
\hline
                              & T5-base & Uniform Weighted & DynaOpt & Model Soups & EMORL (ours) \\ \hline
Diversity-2 ($\color{green}\uparrow$)      & $0.8851 \scriptstyle \pm 0.0056$ & $0.3561 \scriptstyle \pm 0.0837$            & $0.3621 \scriptstyle \pm 0.0951$ & $0.4327 \scriptstyle \pm 0.0932$ & $\mathbf{0.6516} \scriptstyle \pm 0.0524$ \\ 
Edit Rate ($\color{green}\uparrow$)        & $0.8087 \scriptstyle \pm 0.0127$ & $0.8870 \scriptstyle \pm 0.0247$            & $\mathbf{0.8929} \scriptstyle \pm 0.0246$ & $0.8672 \scriptstyle \pm 0.0326$ & $0.8734 \scriptstyle \pm 0.0240$ \\
Data Consumption ($\color{red}\downarrow$)&                    & $23398 \scriptstyle \pm 7390$               & $31328 \scriptstyle \pm 16736$ & $18924 \scriptstyle \pm 4672$ & $\mathbf{17529} \scriptstyle \pm 1650$\\
Time Consumption ($\color{red}\downarrow$)&                    & $5967.84 \scriptstyle \pm 1875.50$ & $8029.15 \scriptstyle \pm 4365.64$ & $\mathbf{5823} \scriptstyle \pm 1262.24$ & $6573 \scriptstyle \pm 147.43$\\ 
Scalability                   & & - & - & + & + \\
Explainability                & & - & - & + & + \\ \hline
\end{tabular}
\label{tab: evaluation metrics}
\end{table*}

\begin{table*}[t]
\centering
\small
\caption{The performance metrics are evaluated automatically and through human assessment on the PAIR and Psych8k datasets. The human-evaluated scores are averaged across 640 samples from both datasets. The overall results demonstrate that our EMORL method achieves performance comparable to other methods. }
\begin{tabular}{cllll}
\hline
 & & Reflection ($\color{green}\uparrow$) & Empathy ($\color{green}\uparrow$) & Fluency ($\color{green}\uparrow$) \\ \hline
\multirow{4}{*}{PAIR} & T5-base & $0.0418 \scriptstyle \pm 0.0108$ & $0.4648 \scriptstyle \pm 0.0160$ & $0.4849 \scriptstyle \pm 0.0185$ \\
 & Uniform Weighted & $\mathbf{0.9616} \scriptstyle \pm 0.0212$ & $0.8078 \scriptstyle \pm 0.0251$ & $\mathbf{0.7498} \scriptstyle \pm 0.0176$ \\ 
 & DynaOpt          & $0.9349 \scriptstyle \pm 0.0234$ & $\mathbf{0.8141} \scriptstyle \pm 0.0329$ & $0.7271 \scriptstyle \pm 0.0300$ \\
 & Model Soups       & $0.9204 \scriptstyle \pm 0.0315$ & $0.7418 \scriptstyle \pm 0.0264$ & $0.4324 \scriptstyle \pm 0.0186$ \\
 & EMORL (ours)     & $0.9406 \scriptstyle \pm 0.0406$ & $0.7766 \scriptstyle \pm 0.0178$ & $0.6548 \scriptstyle \pm 0.0113$ \\ \hline
\multirow{4}{*}{Psych8k} & T5-base & $0.0968 \scriptstyle \pm 0.0099$ & $0.3198 \scriptstyle \pm 0.0129$ & $0.6397 \scriptstyle \pm 0.0062$ \\
 & Uniform Weighted & $0.9694 \scriptstyle \pm 0.0066$ & $0.7317 \scriptstyle \pm 0.0314$ & $\mathbf{0.7897} \scriptstyle \pm 0.0173$ \\
 & DynaOpt          & $0.9755 \scriptstyle \pm 0.0148$ & $\mathbf{0.7330} \scriptstyle \pm 0.0487$ & $0.7725 \scriptstyle \pm 0.0247$ \\ 
 & Model Soups       & $0.9518 \scriptstyle \pm 0.0126$ & $0.6722 \scriptstyle \pm 0.0235$ & $0.4602 \scriptstyle \pm 0.0162$ \\
 & EMORL (ours)     & $\mathbf{0.9784} \scriptstyle \pm 0.0164$ & $0.6838 \scriptstyle \pm 0.0268$ & $0.7462 \scriptstyle \pm 0.0108$ \\ \hline
\multirow{4}{*}{Human} & T5-base & $0.2618$ & $0.2563$ & $0.6875$ \\
 & Uniform Weighted & $0.5074$ & $0.4563$ & $\mathbf{0.4438}$ \\
 & DynaOpt          & $\mathbf{0.5608}$ & $\mathbf{0.5473}$ & $0.3118$ \\ 
 & Model Soups       & $0.5178$ & $0.5122$ & $0.2490$ \\
 & EMORL (ours)     & $0.5308$ & $0.4858$ & $0.3758$ \\ \hline
\end{tabular}
\label{tab: reward metrics}
\end{table*}

\paragraph{Hierarchical Grid Visualization Demonstrates Explainable Results:} Figure~\ref{fig: visualization} illustrates the hierarchical grid search process: in iteration $1$ (left subplot), we evaluated $3\times 3\times 3$ weighted combinations for reflection, empathy, and fluency, with color mapping indicating objective reward values. The $2\times 2\times 2$ area with best total performance was then used to refine the search space for subsequent iterations. The right subplot shows iteration $5$, where the search converged to a more precise and refined space: $(0.75, 0.8125)$, $(0.4375, 0.5)$ and $(0.0, 0.0625)$ for the aggregation weights of reflection, empathy, and fluency models respectively. This progressive refinement allows for the precise identification of the optimal weighted combination.

The visualization reveals significant insights regarding each objective's contribution to overall performance. Unlike defining weight coefficients in the reward function during training, which doesn't correspond linearly to performance, our aggregation phase's contribution distribution accurately reflects each objective-specific model's actual contribution. As shown in Figure~\ref{fig: visualization}, the aggregation benefits (orange points) from higher weights of the reflection model. Empathy delivers optimal overall performance with moderate weights, while extreme weights reduce the objective reward. The fluency model demonstrates a negative effect (cyan points) on other objectives when assigned high weights, with lower weights facilitating better integration with the other objectives. This visualization can be readily acquired during aggregation by evaluating weighted combinations on batches of test data. Although results vary across experiments with different sampled models, as shown in Table~\ref{tab: weights combination} (Section~\ref{sec: appendix}), this approach underscores the interpretability and explainability of EMORL.

\paragraph{EMORL Shows Promise Across Multiple Evaluation Metrics:} As shown in Table~\ref{tab: evaluation metrics}, EMORL achieves the highest diversity-2 score among fine-tuned models, averaging above $0.65$, compared to $0.35-0.43$ for other fine-tuned models. This highlights EMORL's ability to generate diverse responses, which is achieved by aggregating hidden states and delegating the token generation to the language model head. EMORL has an edit rate of $0.8734$, which is very close to other models, indicating that it avoids verbatim repetition.

EMORL demonstrates superior efficiency in resource utilization, consuming approximately $17,529$ data points and $6,573$ seconds of training time. Due to its ensembled architecture, the total resource consumption $T_{total}$ is determined by:
\begin{equation}
    T_{total} = \max_{i \in \{1,2,3\}} \{T_{train}(obj_i)\} + T_{agg}
\end{equation}

This parallel process enables faster training with over $5869$ fewer data points compared to single-policy methods. While another meta-policy method, Model Soups, employs gradient-based algorithms (learned soup) that often struggle with local optima, leading to increased consumption. However, EMORL employs hidden-state level aggregation and requires token-by-token generation, which interrupts the sequential generation process of transformers, slightly increasing time consumption. EMORL exhibits greater stability, with variations of only $1,650$ data points and $147.43$ seconds, significantly lower than DynaOpt's variations of $16,736$ data points and $4,365.64$ seconds. This stability is attributed to EMORL's consistent single-objective fine-tuning and uniform optimization resource consumption facilitated by hierarchical grid search. These advantages position EMORL as an efficient and stable MOO approach.

EMORL is also a flexible fine-tuning approach that achieves both scalability and explainability. Its scalability is demonstrated through two capabilities: 1) Different preferences: EMORL avoids retraining by optimizing weights in the aggregation phase. While our default uses average preferences, user-specific preferences generate new optimal weights during the aggregation phase without costly fine-tuning. 2) New objectives: When incorporating new objectives like helpfulness, EMORL only trains the new objective-specific model and optimizes weights for combining all models, while preserving existing trained models. For explainability, EMORL provides clear insights through hierarchical grid search visualization, as shown in Figure~\ref{fig: visualization}. It also demonstrates how increasing one objective's weight enhances its performance while affecting others. This trade-off is efficiently interpreted through testing only on small data batches. In contrast, conventional single-policy methods require complete retraining for different preferences or new objectives, plus extensive trial-and-error to determine objective importance.

\begin{figure}[t]
\begin{center}
\includegraphics[width=1.0\linewidth]{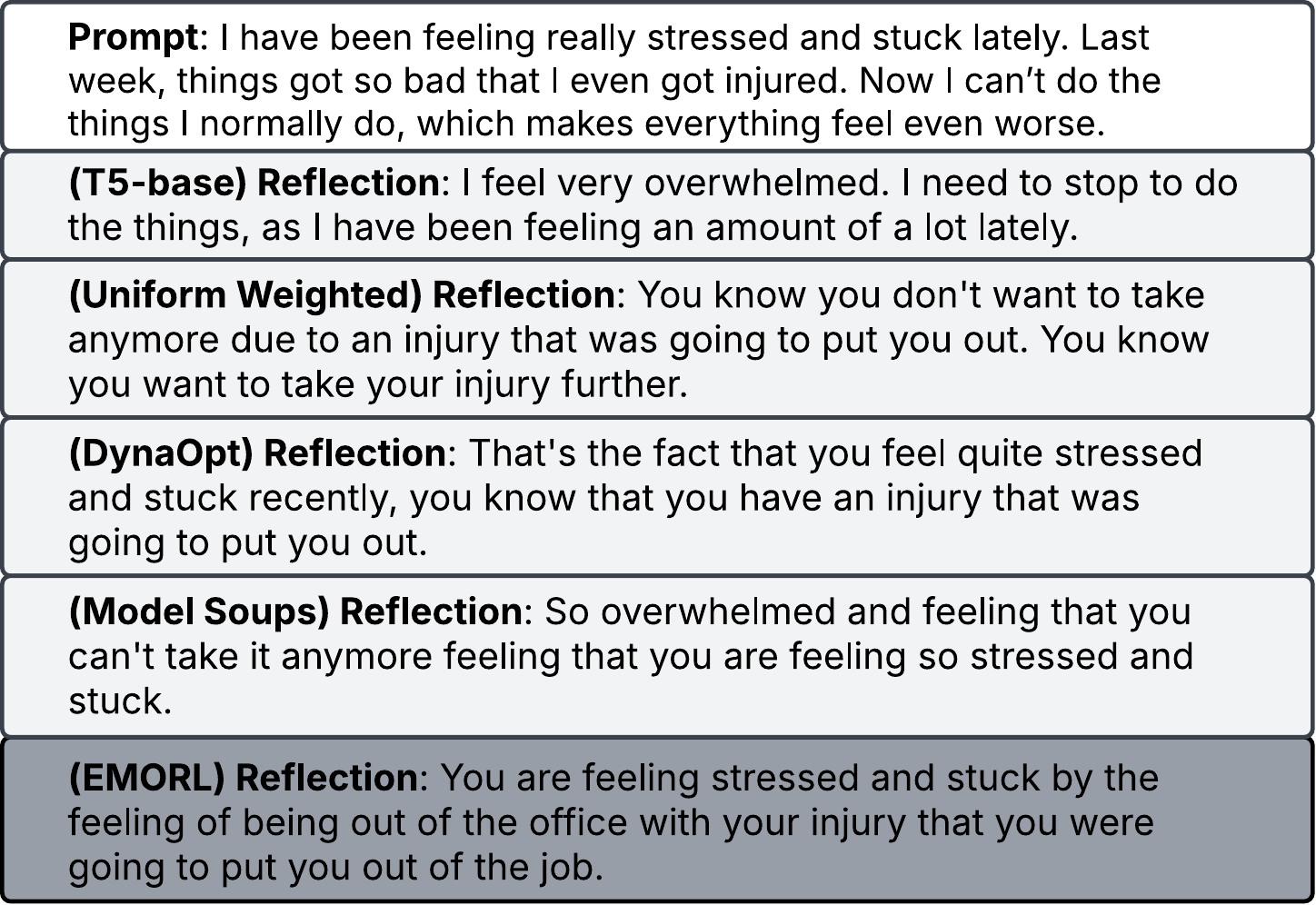}
\end{center}
\caption{Sampled generations of different models on the counselor reflection generation task. }
\label{fig: sample generation}
\end{figure}

\paragraph{EMORL Delivers Comparable Performance in Performance Metrics:} We evaluated our EMORL method on the \emph{PAIR} and \emph{Psych8k} datasets for reflection, empathy, and fluency metrics. Although EMORL does not achieve the highest scores, it maintains performance comparable to the conventional single-policy models and significantly outperforms the Model Soups models.

On the PAIR dataset, EMORL achieves an average score of $0.7907$, outperforming parameter-level aggregation Model Soups ($0.6982$) and performing comparably to the single-policy Uniform Weighted ($0.8397$) and DynaOpt ($0.8254$) methods. Notably, model performance varies slightly between the two datasets. On Psych8k, EMORL achieves the objective reward of $0.8082$, with its reflection score reaching the highest ($0.9784$) among all models. Compared to Model Soups, EMORL shows a substantial improvement of $0.25$ in fluency scores, demonstrating the effectiveness of hidden-state level aggregation for LLM fine-tuning.

Human evaluation scores, averaged across PAIR and Psych8k, are generally lower than automated metrics but show consistent trends. These evaluations support our findings: all fine-tuned models demonstrate improvements in reflection and empathy but exhibit a slight decline in fluency. EMORL achieves the second-highest scores in reflection ($0.5308$), empathy ($0.4858$), and fluency ($0.3758$) among fine-tuned models, demonstrating balanced performance across all metrics and underscoring its potential as an effective fine-tuning method.

The sample generations are shown in Figure~\ref{fig: sample generation}. EMORL improves reflection and empathy by employing second-person speech ("you"), introducing new perspectives ("job", "office"), and crafting understandable and empathetic statements ("feeling") in response to prompts. EMORL can paraphrase prompts by reflecting on the patient's problem and aligning well with the desired objectives for generations, highlighting its effectiveness.

\section{Conclusion}
To conclude, our study addresses the challenges of MOO in LLM fine-tuning. We identify the key limitations of convergence speed, process stability, and performance metrics in conventional single-policy approaches. To address these limitations, we propose EMORL, a novel meta-policy framework using ensemble learning to aggregate diverse models trained on individual objectives. The results demonstrate that EMORL achieves greater diversity, efficiency, scalability, and explainability while maintaining performance comparable to existing methods in counselor generation tasks.

Our approach is the first to aggregate hidden states to incorporate multiple objectives in NLP. Unlike classification or regression models, which can simply concatenate results from each model to collaboratively decide on the final output, LLMs produce complete sequential generations where fluency cannot be separated or neglected. The hidden states represent rich contextual information, which proves more valuable than incorporating parameters or logits. By using the hidden-state level aggregation, EMORL exceeds the performance limitations of meta-policy methods. EMORL also offers unique advantages over conventional single-policy methods, including improved training efficiency, extensive scalability, and enhanced explainability due to the parallel training properties, making it more efficient and flexible. We investigated the nature of the MOO problem and discovered that optima consistently appear within higher-performance regions. This insight led us to design a simple yet effective hierarchical grid search algorithm that requires fewer evaluations to find the globally optimal weights. 

Our approach is both task- and model-agnostic. It's compatible with all transformer-based LLMs, as these architectures maintain hidden states that represent contextual information encompassing multiple objectives in NLP. This study demonstrates the potential of ensemble learning to advance current RL training paradigms and points to a promising novel direction for efficient and flexible MOO in future applications.




\section*{Limitations}
Our study has advanced a new paradigm of multi-objective optimization for LLM fine-tuning but faces several limitations that suggest directions for future research. First, the current implementation focuses on single-turn generation, which fails to capture the dynamics of counseling conversations. The RL interaction is limited to one-time evaluations without dialogue history, and reflections are generated based solely on prompts, not fully leveraging RL's potential for complex interactions. Future work should explore multi-turn conversation tasks, potentially incorporating dynamic weighting of model aggregation across dialogue turns.

Second, our study employs small-scale encoder-decoder LLMs, which may not achieve application-level performance. As shown in Figure~\ref{fig: sample generation}, while EMORL generates the reflections addressing new empathic perspectives compared to the pre-trained model, the overall quality remains limited. This indicates that pre-trained model constraints affect generation quality, despite improvements in targeted behaviors. Future research should implement EMORL on larger models with billions of parameters to enhance performance and capabilities.

Finally, challenges remain in effectiveness and efficiency. While EMORL achieves comparable results across objectives, improving performance to surpass conventional RL fine-tuning remains a key challenge, which could potentially be addressed through non-linear combination approaches. Additionally, hidden-state level aggregation requires token-by-token generation, impacting the sequential generation process and slightly increasing time consumption. Future work should explore advanced aggregation methods to enhance computational efficiency and output quality while preserving the benefits of ensemble learning.

\section*{Potential Risks}
We suggest that our models are not advocated for deployment in clinical or mental health settings. This is because human understanding and communication are indispensable in these domains, and the behavior of language models remains incompletely explored. Instead, we propose that our method and models be utilized for methodological research. In future real-world applications, we recommend making users explicitly aware that they are communicating with AI-generated content that does not represent genuine human intelligence or empathy, thereby mitigating ELIZA effects where users may incorrectly attribute human-like qualities to AI systems.

\subsection*{Ethical Considerations}
The PAIR and Psych8K datasets used in our study are either open-source or licensed under CC-BY-NC. These datasets include one-turn motivational interviewing conversations as well as mental health interactions between counselors and patients. We ensured that the source datasets processed the dialogues to redact any personally identifiable information. Generative AI was employed solely to assist with bug fixing and grammatical error correction. All other work presented in this paper was conducted entirely by us.

\section*{Acknowledgements}
This work has been supported by the ARGUS EU project (Grant Agreement No. 101132308), funded by the European Union. Views and opinions expressed are however those of the author(s) only and do not necessarily reflect those of the European Union or of the European Research Executive Agency (REA). Neither the European Union nor the granting authority can be held responsible for them. The contribution of Cong Yang was supported by the National Natural Science Foundation of China (Grant No. 62473276), the Natural Science Foundation of Jiangsu Province (Grant No. BK20241918), and the Research Fund of Horizon Robotics (Grant No. H230666). We thank Yaoting Zhu from Zhejiang Sci-Tech University for her support on human evaluation.

\bibliography{main}

\begin{thebibliography}{30}
\providecommand{\natexlab}[1]{#1}

\bibitem[{Bishop and Nasrabadi(2007)}]{bishop2006pattern}
Christopher~M. Bishop and Nasser~M. Nasrabadi. 2007.
\newblock \href {https://doi.org/10.1117/1.2819119} {\emph{Pattern Recognition and Machine Learning}}.
\newblock \emph{J. Electronic Imaging}, 16(4):049901.

\bibitem[{Dann et~al.(2023)Dann, Mansour, and Mohri}]{dann2023reinforcement}
Christoph Dann, Yishay Mansour, and Mehryar Mohri. 2023.
\newblock \href {https://proceedings.mlr.press/v202/dann23a.html} {Reinforcement learning can be more efficient with multiple rewards}.
\newblock In \emph{International Conference on Machine Learning, {ICML} 2023, 23-29 July 2023, Honolulu, Hawaii, {USA}}, volume 202 of \emph{Proceedings of Machine Learning Research}, pages 6948--6967. {PMLR}.

\bibitem[{Dulac{-}Arnold et~al.(2021)Dulac{-}Arnold, Levine, Mankowitz, Li, Paduraru, Gowal, and Hester}]{dulac2021challenges}
Gabriel Dulac{-}Arnold, Nir Levine, Daniel~J. Mankowitz, Jerry Li, Cosmin Paduraru, Sven Gowal, and Todd Hester. 2021.
\newblock \href {https://doi.org/10.1007/s10994-021-05961-4} {Challenges of real-world reinforcement learning: definitions, benchmarks and analysis}.
\newblock \emph{Machine Learning}, 110(9):2419--2468.

\bibitem[{Feriani et~al.(2022)Feriani, Wu, Xu, Li, Jang, Hossain, Liu, and Dudek}]{feriani2022multiobjective}
Amal Feriani, Di~Wu, Yi~Tian Xu, Jimmy Li, Seowoo Jang, Ekram Hossain, Xue Liu, and Gregory Dudek. 2022.
\newblock \href {https://doi.org/10.1109/JSAC.2022.3191114} {Multiobjective load balancing for multiband downlink cellular networks: {A} meta- reinforcement learning approach}.
\newblock \emph{{IEEE} J. Sel. Areas Commun.}, 40(9):2614--2629.

\bibitem[{Gautschi(2010)}]{gautschi2011numerical}
Walter Gautschi. 2010.
\newblock \href {https://doi.org/10.1016/J.CAM.2009.11.054} {The spiral of theodorus, numerical analysis, and special functions}.
\newblock \emph{J. Comput. Appl. Math.}, 235(4):1042--1052.

\bibitem[{Hayes et~al.(2022)Hayes, Radulescu, Bargiacchi, K{\"{a}}llstr{\"{o}}m, Macfarlane, Reymond, Verstraeten, Zintgraf, Dazeley, Heintz, Howley, Irissappane, Mannion, Now{\'{e}}, de~Oliveira~Ramos, Restelli, Vamplew, and Roijers}]{hayes2022practical}
Conor~F. Hayes, Roxana Radulescu, Eugenio Bargiacchi, Johan K{\"{a}}llstr{\"{o}}m, Matthew Macfarlane, Mathieu Reymond, Timothy Verstraeten, Luisa~M. Zintgraf, Richard Dazeley, Fredrik Heintz, Enda Howley, Athirai~A. Irissappane, Patrick Mannion, Ann Now{\'{e}}, Gabriel de~Oliveira~Ramos, Marcello Restelli, Peter Vamplew, and Diederik~M. Roijers. 2022.
\newblock \href {https://doi.org/10.1007/S10458-022-09552-Y} {A practical guide to multi-objective reinforcement learning and planning}.
\newblock \emph{Auton. Agents Multi Agent Syst.}, 36(1):26.

\bibitem[{Hu et~al.(2022)Hu, Shen, Wallis, Allen{-}Zhu, Li, Wang, Wang, and Chen}]{hu2021lora}
Edward~J. Hu, Yelong Shen, Phillip Wallis, Zeyuan Allen{-}Zhu, Yuanzhi Li, Shean Wang, Lu~Wang, and Weizhu Chen. 2022.
\newblock \href {https://openreview.net/forum?id=nZeVKeeFYf9} {Lora: Low-rank adaptation of large language models}.
\newblock In \emph{The Tenth International Conference on Learning Representations, {ICLR} 2022, Virtual Event, April 25-29, 2022}. OpenReview.net.

\bibitem[{Kumar(2024)}]{kumar2024large}
Pranjal Kumar. 2024.
\newblock \href {https://doi.org/10.1007/S10462-024-10888-Y} {Large language models (llms): survey, technical frameworks, and future challenges}.
\newblock \emph{Artif. Intell. Rev.}, 57(9):260.

\bibitem[{Laban et~al.(2021)Laban, Schnabel, Bennett, and Hearst}]{laban2021keep}
Philippe Laban, Tobias Schnabel, Paul~N. Bennett, and Marti~A. Hearst. 2021.
\newblock \href {https://doi.org/10.18653/V1/2021.ACL-LONG.498} {Keep it simple: Unsupervised simplification of multi-paragraph text}.
\newblock In \emph{Proceedings of the 59th Annual Meeting of the Association for Computational Linguistics and the 11th International Joint Conference on Natural Language Processing, {ACL/IJCNLP} 2021, (Volume 1: Long Papers), Virtual Event, August 1-6, 2021}, pages 6365--6378. Association for Computational Linguistics.

\bibitem[{Lin(2024)}]{lin2024reinforcement}
Baihan Lin. 2024.
\newblock \href {https://doi.org/10.1007/978-3-031-53720-2_15} {Reinforcement learning in large language models (llms): The rise of {AI} language giants}.
\newblock In \emph{Reinforcement Learning Methods in Speech and Language Technology}, pages 147--156. Springer.

\bibitem[{Malik et~al.(2024)Malik, Iqbal, Sharif, Shah, Khalil, Irfan, and Rosak{-}Szyrocka}]{malik2024attention}
Shanza~Zafar Malik, Khalid Iqbal, Muhammad Sharif, Yaser~Ali Shah, Amaad Khalil, Muhammad~Abeer Irfan, and Joanna Rosak{-}Szyrocka. 2024.
\newblock \href {https://doi.org/10.7717/PEERJ-CS.2283} {Attention-aware with stacked embedding for sentiment analysis of student feedback through deep learning techniques}.
\newblock \emph{PeerJ Comput. Sci.}, 10:e2283.

\bibitem[{Min et~al.(2024)Min, P{\'{e}}rez{-}Rosas, Resnicow, and Mihalcea}]{perez2024dynamic}
Do~June Min, Ver{\'{o}}nica P{\'{e}}rez{-}Rosas, Ken Resnicow, and Rada Mihalcea. 2024.
\newblock \href {https://aclanthology.org/2024.lrec-main.483} {Dynamic reward adjustment in multi-reward reinforcement learning for counselor reflection generation}.
\newblock In \emph{Proceedings of the 2024 Joint International Conference on Computational Linguistics, Language Resources and Evaluation, {LREC/COLING} 2024, 20-25 May, 2024, Torino, Italy}, pages 5437--5449. {ELRA} and {ICCL}.

\bibitem[{Min et~al.(2022)Min, P{\'{e}}rez{-}Rosas, Resnicow, and Mihalcea}]{min-etal-2022-pair}
Do~June Min, Ver{\'{o}}nica P{\'{e}}rez{-}Rosas, Kenneth Resnicow, and Rada Mihalcea. 2022.
\newblock \href {https://doi.org/10.18653/V1/2022.EMNLP-MAIN.11} {{PAIR:} prompt-aware margin ranking for counselor reflection scoring in motivational interviewing}.
\newblock In \emph{Proceedings of the 2022 Conference on Empirical Methods in Natural Language Processing, {EMNLP} 2022, Abu Dhabi, United Arab Emirates, December 7-11, 2022}, pages 148--158. Association for Computational Linguistics.

\bibitem[{Mohan et~al.(2023)Mohan, Benjamins, Wienecke, Dockhorn, and Lindauer}]{mohan2023autorl}
Aditya Mohan, Carolin Benjamins, Konrad Wienecke, Alexander Dockhorn, and Marius Lindauer. 2023.
\newblock \href {https://proceedings.mlr.press/v224/mohan23a.html} {Autorl hyperparameter landscapes}.
\newblock In \emph{International Conference on Automated Machine Learning, 12-15 November 2023, Hasso Plattner Institute, Potsdam, Germany}, volume 224 of \emph{Proceedings of Machine Learning Research}, pages 13/1--27. {PMLR}.

\bibitem[{Okano et~al.(2023)Okano, Funakoshi, Nagata, and Okumura}]{okano2023generating}
Yuki Okano, Kotaro Funakoshi, Ryo Nagata, and Manabu Okumura. 2023.
\newblock \href {https://doi.org/10.18653/V1/2023.BEA-1.16} {Generating dialog responses with specified grammatical items for second language learning}.
\newblock In \emph{Proceedings of the 18th Workshop on Innovative Use of {NLP} for Building Educational Applications, BEA@ACL 2023, Toronto, Canada, 13 July 2023}, pages 184--194. Association for Computational Linguistics.

\bibitem[{O'neil et~al.(2023)O'neil, Sedoc, Yang, Zhu, and Ungar}]{o2023automatic}
Emma O'neil, Jo{\~a}o Sedoc, Diyi Yang, Haiyi Zhu, and Lyle Ungar. 2023.
\newblock \href {https://aclanthology.org/2023.gem-1.6/} {Automatic reflection generation for peer-to-peer counseling}.
\newblock In \emph{Proceedings of the Third Workshop on Natural Language Generation, Evaluation, and Metrics ({GEM})}, pages 62--75.

\bibitem[{Ouyang et~al.(2022)Ouyang, Wu, Jiang, Almeida, Wainwright, Mishkin, Zhang, Agarwal, Slama, Ray, Schulman, Hilton, Kelton, Miller, Simens, Askell, Welinder, Christiano, Leike, and Lowe}]{ouyang2022training}
Long Ouyang, Jeffrey Wu, Xu~Jiang, Diogo Almeida, Carroll~L. Wainwright, Pamela Mishkin, Chong Zhang, Sandhini Agarwal, Katarina Slama, Alex Ray, John Schulman, Jacob Hilton, Fraser Kelton, Luke Miller, Maddie Simens, Amanda Askell, Peter Welinder, Paul~F. Christiano, Jan Leike, and Ryan Lowe. 2022.
\newblock \href {http://papers.nips.cc/paper\_files/paper/2022/hash/b1efde53be364a73914f58805a001731-Abstract-Conference.html} {Training language models to follow instructions with human feedback}.
\newblock In \emph{Advances in Neural Information Processing Systems 35: Annual Conference on Neural Information Processing Systems 2022, NeurIPS 2022, New Orleans, LA, USA, November 28 - December 9, 2022}.

\bibitem[{Parthasarathy et~al.(2024)Parthasarathy, Zafar, khan, and Shahid}]{parthasarathy2024ultimate}
Venkatesh~Balavadhani Parthasarathy, Ahtsham Zafar, Aafaq~Iqbal khan, and Arsalan Shahid. 2024.
\newblock \href {https://doi.org/10.48550/ARXIV.2408.13296} {The ultimate guide to fine-tuning llms from basics to breakthroughs: An exhaustive review of technologies, research, best practices, applied research challenges and opportunities}.
\newblock \emph{CoRR}, abs/2408.13296.

\bibitem[{Pu et~al.(2024)Pu, Fu, Dong, Zhang, and Liu}]{pu2024dynamic}
Juncheng Pu, Xiaodong Fu, Hai Dong, Pengcheng Zhang, and Li~Liu. 2024.
\newblock \href {https://doi.org/10.1109/TSC.2024.3478796} {Dynamic adaptive federated learning on local long-tailed data}.
\newblock \emph{{IEEE} Trans. Serv. Comput.}, 17(6):3485--3498.

\bibitem[{Qin et~al.(2024)Qin, Chen, Feng, Wu, Zhang, Li, Li, Che, and Yu}]{qin2024large}
Libo Qin, Qiguang Chen, Xiachong Feng, Yang Wu, Yongheng Zhang, Yinghui Li, Min Li, Wanxiang Che, and Philip~S. Yu. 2024.
\newblock \href {https://doi.org/10.48550/ARXIV.2405.12819} {Large language models meet {NLP:} {A} survey}.
\newblock \emph{CoRR}, abs/2405.12819.

\bibitem[{Raffel et~al.(2020)Raffel, Shazeer, Roberts, Lee, Narang, Matena, Zhou, Li, and Liu}]{raffel2020exploring}
Colin Raffel, Noam Shazeer, Adam Roberts, Katherine Lee, Sharan Narang, Michael Matena, Yanqi Zhou, Wei Li, and Peter~J. Liu. 2020.
\newblock \href {https://jmlr.org/papers/v21/20-074.html} {Exploring the limits of transfer learning with a unified text-to-text transformer}.
\newblock \emph{J. Mach. Learn. Res.}, 21:140:1--140:67.

\bibitem[{Shi et~al.(2024)Shi, Chen, Hu, Liu, Hajishirzi, Smith, and Du}]{shi2024decoding}
Ruizhe Shi, Yifang Chen, Yushi Hu, Alisa Liu, Hanna Hajishirzi, Noah~A. Smith, and Simon~S. Du. 2024.
\newblock \href {http://papers.nips.cc/paper\_files/paper/2024/hash/57c89126d60c209f48d0e6395c766bb3-Abstract-Conference.html} {Decoding-time language model alignment with multiple objectives}.
\newblock In \emph{Advances in Neural Information Processing Systems 38: Annual Conference on Neural Information Processing Systems 2024, NeurIPS 2024, Vancouver, BC, Canada, December 10 - 15, 2024}.

\bibitem[{Song et~al.(2024)Song, Suganthan, Pedrycz, Ou, He, Chen, and Wu}]{song2023ensemble}
Yanjie Song, Ponnuthurai~Nagaratnam Suganthan, Witold Pedrycz, Junwei Ou, Yongming He, Ying{-}Wu Chen, and Yutong Wu. 2024.
\newblock \href {https://doi.org/10.1016/J.ASOC.2023.110975} {Ensemble reinforcement learning: {A} survey}.
\newblock \emph{Appl. Soft Comput.}, 149(Part {A}):110975.

\bibitem[{Vamplew et~al.(2022)Vamplew, Smith, K{\"{a}}llstr{\"{o}}m, de~Oliveira~Ramos, Radulescu, Roijers, Hayes, Heintz, Mannion, Libin, Dazeley, and Foale}]{vamplew2022scalar}
Peter Vamplew, Benjamin~J. Smith, Johan K{\"{a}}llstr{\"{o}}m, Gabriel de~Oliveira~Ramos, Roxana Radulescu, Diederik~M. Roijers, Conor~F. Hayes, Fredrik Heintz, Patrick Mannion, Pieter J.~K. Libin, Richard Dazeley, and Cameron Foale. 2022.
\newblock \href {https://doi.org/10.1007/S10458-022-09575-5} {Scalar reward is not enough: a response to silver, singh, precup and sutton {(2021)}}.
\newblock \emph{Auton. Agents Multi Agent Syst.}, 36(2):41.

\bibitem[{Vanhaesebrouck et~al.(2017)Vanhaesebrouck, Bellet, and Tommasi}]{vanhaesebrouck2017decentralized}
Paul Vanhaesebrouck, Aur{\'{e}}lien Bellet, and Marc Tommasi. 2017.
\newblock \href {http://proceedings.mlr.press/v54/vanhaesebrouck17a.html} {Decentralized collaborative learning of personalized models over networks}.
\newblock In \emph{Proceedings of the 20th International Conference on Artificial Intelligence and Statistics, {AISTATS} 2017, 20-22 April 2017, Fort Lauderdale, FL, {USA}}, volume~54 of \emph{Proceedings of Machine Learning Research}, pages 509--517. {PMLR}.

\bibitem[{Vaswani et~al.(2017)Vaswani, Shazeer, Parmar, Uszkoreit, Jones, Gomez, Kaiser, and Polosukhin}]{vaswani2017attention}
Ashish Vaswani, Noam Shazeer, Niki Parmar, Jakob Uszkoreit, Llion Jones, Aidan~N. Gomez, Lukasz Kaiser, and Illia Polosukhin. 2017.
\newblock \href {https://proceedings.neurips.cc/paper/2017/hash/3f5ee243547dee91fbd053c1c4a845aa-Abstract.html} {Attention is all you need}.
\newblock In \emph{Advances in Neural Information Processing Systems 30: Annual Conference on Neural Information Processing Systems 2017, December 4-9, 2017, Long Beach, CA, {USA}}, pages 5998--6008.

\bibitem[{Wortsman et~al.(2022)Wortsman, Ilharco, Gadre, Roelofs, Lopes, Morcos, Namkoong, Farhadi, Carmon, Kornblith, and Schmidt}]{wortsman2022model}
Mitchell Wortsman, Gabriel Ilharco, Samir~Yitzhak Gadre, Rebecca Roelofs, Raphael~Gontijo Lopes, Ari~S. Morcos, Hongseok Namkoong, Ali Farhadi, Yair Carmon, Simon Kornblith, and Ludwig Schmidt. 2022.
\newblock \href {https://proceedings.mlr.press/v162/wortsman22a.html} {Model soups: averaging weights of multiple fine-tuned models improves accuracy without increasing inference time}.
\newblock In \emph{International Conference on Machine Learning, {ICML} 2022, 17-23 July 2022, Baltimore, Maryland, {USA}}, volume 162 of \emph{Proceedings of Machine Learning Research}, pages 23965--23998. {PMLR}.

\bibitem[{Yuan and Lai(2024)}]{yuan2025towards}
Qin Yuan and Yuping Lai. 2024.
\newblock \href {https://doi.org/10.1007/s40031-024-01178-w} {Towards efficient information retrieval in {Internet} of {Things} environments via machine learning approaches}.
\newblock \emph{Journal of The Institution of Engineers (India): Series {B}}, 106(1):363--386.

\bibitem[{Zhang et~al.(2023)Zhang, Wu, Zhang, and Wang}]{zhang2022meta}
Zizhen Zhang, Zhiyuan Wu, Hang Zhang, and Jiahai Wang. 2023.
\newblock \href {https://doi.org/10.1109/TNNLS.2022.3148435} {Meta-learning-based deep reinforcement learning for multiobjective optimization problems}.
\newblock \emph{{IEEE} Trans. Neural Networks Learn. Syst.}, 34(10):7978--7991.

\bibitem[{Ziegler et~al.(2019)Ziegler, Stiennon, Wu, Brown, Radford, Amodei, Christiano, and Irving}]{ziegler2019fine}
Daniel~M. Ziegler, Nisan Stiennon, Jeffrey Wu, Tom~B. Brown, Alec Radford, Dario Amodei, Paul~F. Christiano, and Geoffrey Irving. 2019.
\newblock \href {https://arxiv.org/abs/1909.08593} {Fine-tuning language models from human preferences}.
\newblock \emph{CoRR}, abs/1909.08593.

\end{thebibliography}

\appendix
\section{Appendix}
\label{sec: appendix}

\begin{table*}[t]
\centering
\small
\caption{Comparison of single-objective and multi-objective fine-tuning in addressing the challenges. }
\vspace{-1em}
\begin{tabular}{llll} 
\hline
 & Objective Reward ($\color{green}\uparrow$)& Data Consumption ($\color{red}\downarrow$)& Time Consumption ($\color{red}\downarrow$)\\ \hline 
Reflection & $0.9967\pm 0.0028$ & $13209\pm 335$& $4175.05\pm 104.40$\\ 
Empathy    & $0.9935\pm 0.0037$ & $7136\pm 474$& $2232.18\pm 82.62$\\ 
Fluency    & $0.8803\pm 0.0003$ & $4809\pm 178$& $1629.19\pm 58.03$\\ \hline
Uniform Weighted         & $0.8489\pm 0.0172$ & $23398\pm 7390$ & $5967.84\pm 1875.50$ \\ 
DynaOpt                  & $0.8318\pm 0.0076$ & $31328\pm 16736$ & $8029.15\pm 4365.64$ \\ \hline
\end{tabular}
\vspace{-1em}
\label{tab: fine-tuning detail}
\end{table*}

\begin{table}[t]
\centering
\small
\caption{PAIR and Psych8k datasets statistics. }
\vspace{-1em}
\begin{tabular}{lcc}
\hline
statistics & PAIR dataset & Psych8k dataset \\ \hline
\# of Exchange Pairs & 2,544 & 8,187 \\
Avg \# of Words  & 32.39 & 45.18 \\ \hline
\end{tabular}
\vspace{-1em}
\label{tab: datasets statistics}
\end{table}

\begin{table}[t]
\centering
\small
\caption{Demonstration of EMORL's optimal weight combinations for 5 model pairs with different seeds in the experiments, along with their average combination. }
\vspace{-1em}
\begin{tabular}{cllll}
\hline
 & \textbf{Rewards} & Reflection & Empathy & Fluency \\ \hline
$A_1$ & \textbf{0.7936} & 0.9375  & 0.71875 & 0.0625 \\
$A_2$ & \textbf{0.7960} & 1.0     & 0.625   & 0.125  \\
$A_3$ & \textbf{0.8092} & 1.0     & 0.625   & 0.125  \\
$A_4$ & \textbf{0.7942} & 1.0     & 0.5     & 0.0625 \\
$A_5$ & \textbf{0.8093} & 0.78125 & 0.5     & 0.0625 \\ \hline
$\bar{A}$ & \textbf{0.8005} & 0.94375 & 0.59365 & 0.0875 \\ \hline
\end{tabular}
\vspace{-1em}
\label{tab: weights combination}
\end{table}

\begin{figure}[ht]
\begin{center}
\includegraphics[width=0.94\linewidth]{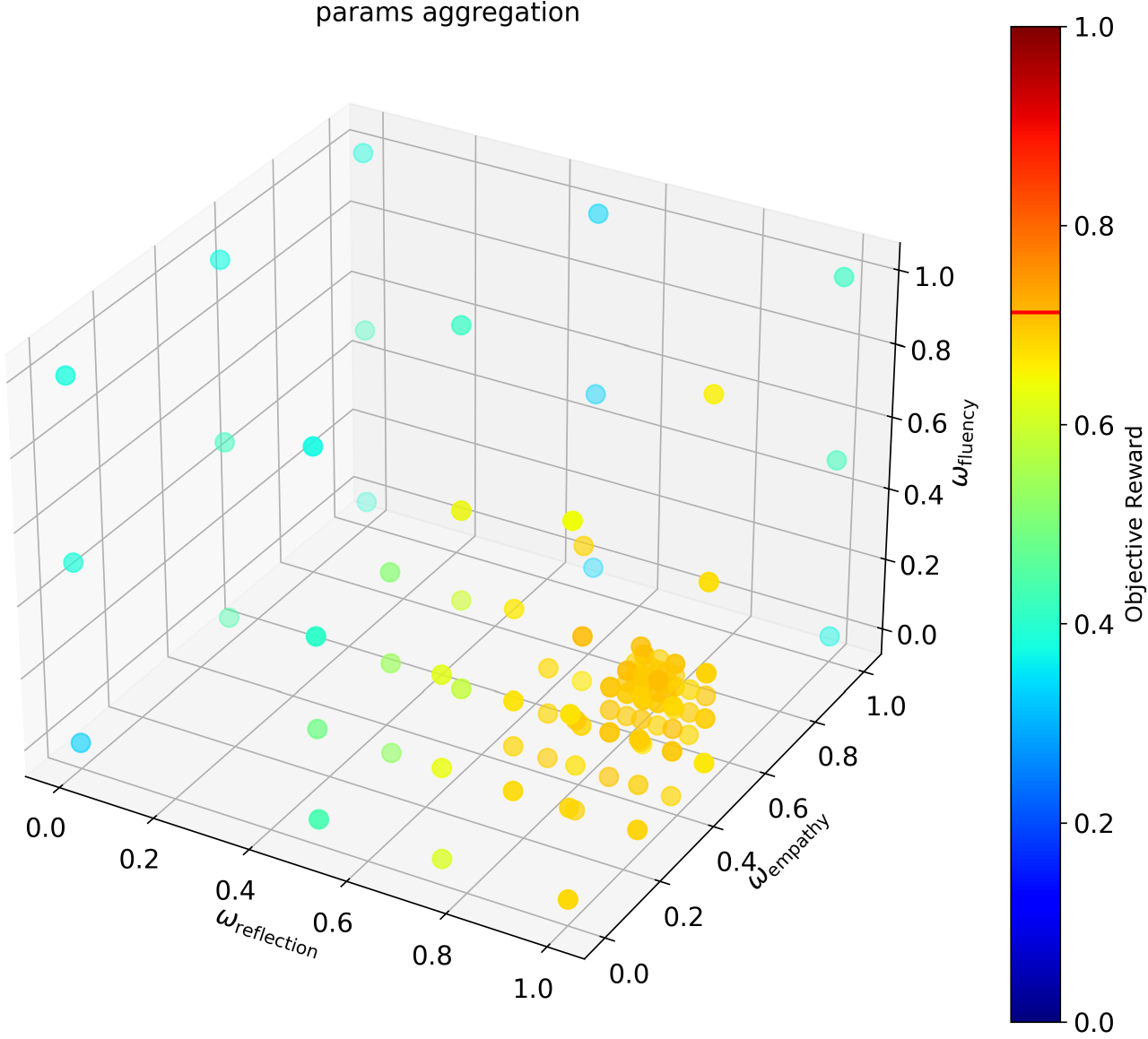}
\end{center}
\vspace{-1em}
\caption{Parameter-level aggregation results. }
\vspace{-1em}
\label{fig: params visualization}
\end{figure}

\begin{figure}[ht]
\begin{center}
\includegraphics[width=0.94\linewidth]{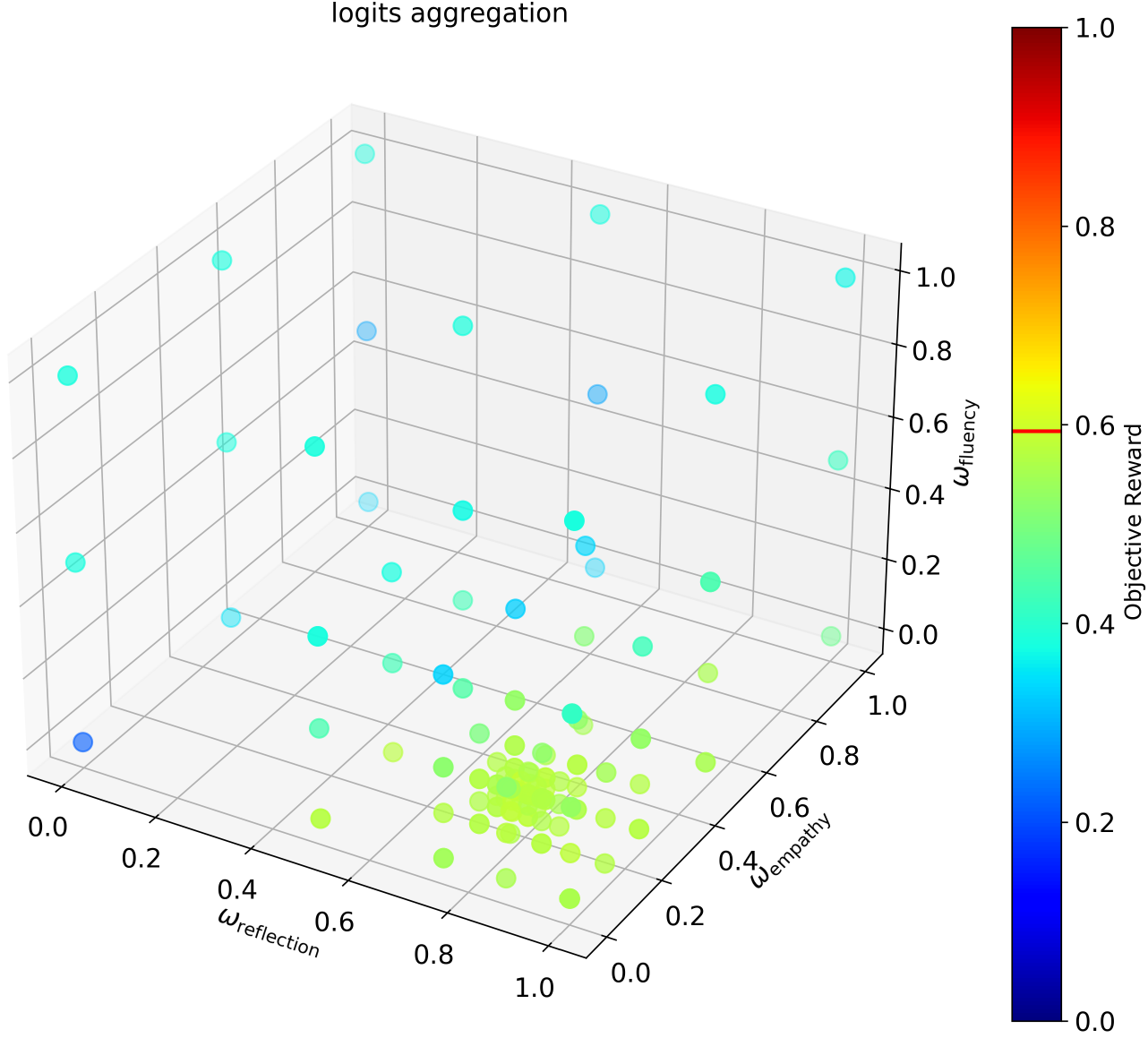}
\end{center}
\vspace{-1em}
\caption{Logit-level aggregation results. }
\vspace{-1em}
\label{fig: logits visualization}
\end{figure}

\subsection{Parameter-level aggregation}
\label{subsec: params agg}

We investigated parameter-level aggregation of local models using LoRA updates, following Equation~\eqref{eq: lora_update}, where the final parameter matrix $\theta$ is constructed by adding weighted low-rank adaptations to the initial pre-trained parameters $\theta_0$. Each adaptation consists of a pair of matrices $B_i$ and $A_i$ whose product forms a low-rank update, scaled by a scaling factor $\alpha_i$, thus avoiding high-rank parameter updates. It focuses on optimizing the weights $w_i$ to effectively incorporate each objective in the aggregation process.

\begin{equation}
    \theta = \theta_0 + \sum_{i=1}^{n} (B_i A_i) \alpha_i w_i \label{eq: lora_update}
\end{equation}

Model Soups~\citep{wortsman2022model}, a prominent ensemble learning method, utilizes this parameter-level aggregation strategy for optimizing hyperparameter configurations. However, when applied to MOO, this approach yielded suboptimal results. As illustrated in Figure~\ref{fig: params visualization}, the overall objective reward achieved only $0.6982$, significantly lower than our hidden-state level aggregation method. The fluency metric performed particularly poorly, reaching merely $0.4324$. This underperformance likely stems from the fundamental incompatibility between diverse model objectives at the parameter-level, ultimately failing to effectively combine multiple optimization objectives.

\subsection{Logit-level aggregation}
\label{subsec: logits agg}
We further explored logit-level aggregation as an alternative approach, which is grounded in the work of \cite{shi2024decoding}. This method aggregates the logits, which represent token probabilities across the vocabulary and directly influence token generation. As illustrated in Figure~\ref{fig: logits visualization}, this approach performed even worse for combining multiple objectives, achieving a maximum objective reward of only $0.5934$, with the fluency metric scoring a mere $0.1575$. This poor performance can be attributed to the naive combination of vocabulary probabilities, where predicted token distributions from different local models are simply merged. This process severely impacts fluency, as the resulting text lacks coherence when calculated from disconnected probability distributions. In contrast, our hidden states aggregation proves more effective by preserving and combining high-level contextual representations during generation, maintaining semantic consistency while still incorporating multiple objectives.

\subsection{Optimization Algorithms}
\label{subsec: bayesian optimization}

\begin{figure}[t]
\begin{center}
\includegraphics[width=0.95\linewidth]{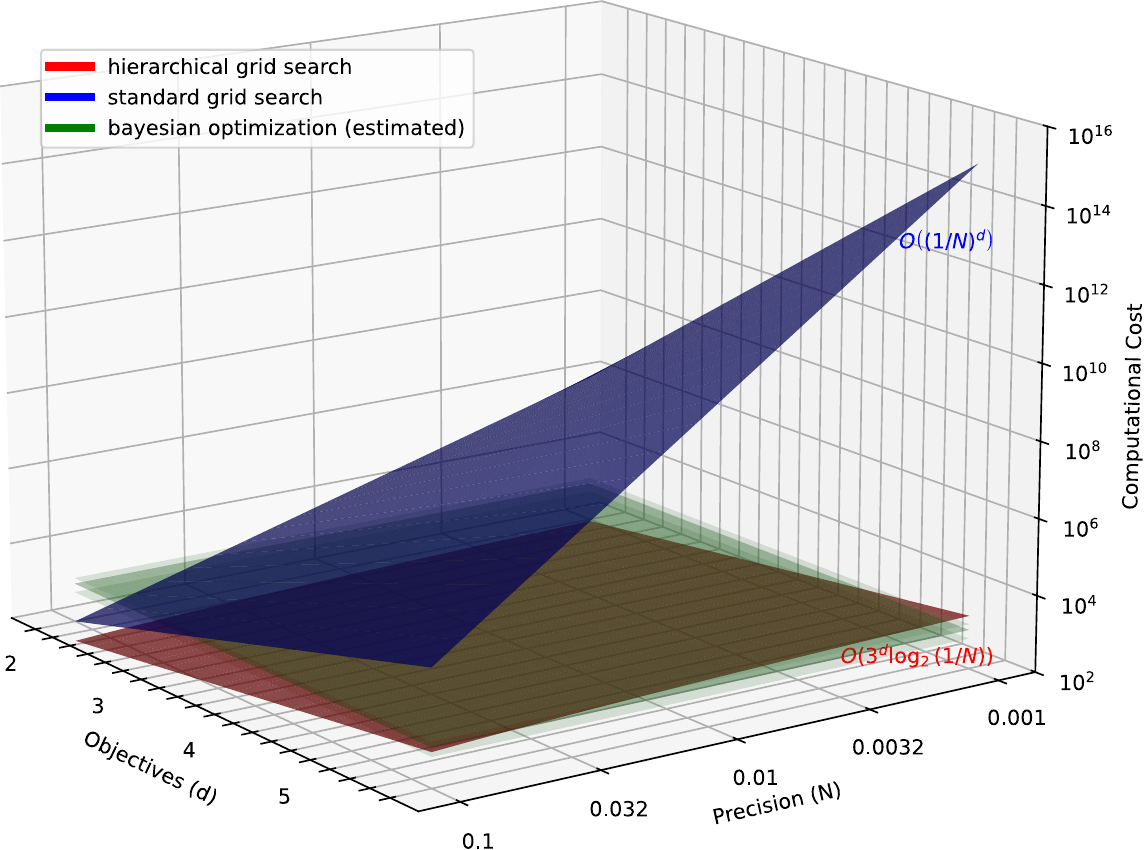}
\end{center}
\vspace{-1em}
\caption{Computational complexity comparison between hierarchical grid search, standard grid search, and Bayesian optimization across varying numbers of objectives and precision levels. }
\vspace{-1em}
\label{fig: complexity}
\end{figure}

For a $d$-dimensional standard grid search with precision level $N$, the computational complexity is $O(\frac{1}{N}^d)$. This follows directly from evaluating the function at $\frac{1}{N}$ grid points in each dimension. In contrast, hierarchical grid search achieves $O(3^d\cdot log_2\frac{1}{N})$ complexity by employing a hierarchical strategy. Starting with a coarse grid, the algorithm progressively refines only promising regions, halving the grid spacing at each level. To reach a final precision of $N$, it requires $log_2\frac{1}{N}$ refinement levels. At each level, we evaluate $3^d$ points per promising region. While still exponential in dimension, the logarithmic dependence on precision offers substantial computational savings for fine-grained searches. A computational complexity comparison among these methods is shown in Figure~\ref{fig: complexity}, where the cost of Bayesian optimization is derived from our experimental estimates.

Bayesian optimization is a sequential strategy for optimizing black-box functions that are expensive to evaluate. Its process consists of building a probabilistic model (e.g., Gaussian Process) of the objective function based on previous observations, using this model to construct an acquisition function that determines where to sample next, evaluating the true objective function at this new point, updating the probabilistic model with this new observation, and repeating until convergence.

We applied Bayesian optimization to our weights optimization problem by allowing it to select appropriate weighted combinations and evaluate their generation performance. We conducted the optimization experiment across several runs. The results showed that in approximately $\mathbf{1034 \pm 423}$ evaluations, Bayesian optimization could identify weights similar to those obtained through hierarchical grid search. Although the weights are continuous without specific precision levels, Bayesian optimization required significantly more computational resources compared to our hierarchical grid search, which evaluated only $\mathbf{135}$ combinations with a precision level of $0.03125$. Furthermore, the stochastic nature of Bayesian optimization led to considerable instability in the optimization process.

We analyze the reason behind why hierarchical grid search is much more effective in this weights optimization problem. The hierarchical approach succeeds primarily because it exploits the structure of the weight space by systematically narrowing down promising regions. Rather than treating the objective function as a complete black box like Bayesian optimization does, hierarchical grid search leverages our prior knowledge that optimal weights likely exist within certain bounded regions. This allows for efficient pruning of unpromising areas early in the search process. Additionally, the deterministic nature of grid search provides consistent, reproducible results, eliminating the variability introduced by the stochastic nature of Bayesian methods. The precision level we implemented ($0.03125$) also worked well for this specific application, as this granularity proved sufficient for practical performance while dramatically reducing the search space compared to the continuous optimization attempted by Bayesian methods. Furthermore, the computational overhead of maintaining and updating probabilistic models in Bayesian optimization becomes significant when the objective function itself is relatively inexpensive to evaluate, making the more straightforward grid-based approach more efficient in this task.

\subsection{Evaluation Instruction}
\label{subsec: instruction}
The human evaluation is supported by two annotators, one is from China, and the other is from Germany. The evaluation, based on their cross-cultural understanding, supports the robust human-annotated results. This evaluation instruction is derived from the original instruction presented in the work~\cite{perez2024dynamic}. When evaluating responses, choose the most appropriate score (0, 1, or 2) based on these criteria. Responses may vary in complexity, and the judgment should be guided by the degree to which they reflect upon the client’s prompt.

\textbf{Reflection}: 0 (Non-Reflection), 1( Simple Reflection), or 2 (Complex Reflection). Non-Reflection (0): A response is considered a non-reflection when it does not engage with the client’s input or the task at hand. It may be off-topic, irrelevant, or simply fail to address the client’s query. Simple Reflection (1): A response is categorized as a simple reflection when it acknowledges the client’s input or question without adding substantial depth or insight. It might repeat or rephrase the client’s words, showing understanding but not extending the conversation significantly. Simple reflections demonstrate basic engagement with the client’squery. Complex Reflection (2): A response is identified as a complex reflection when it goes beyond mere acknowledgment and engages deeply with the client’s input or question. It demonstrates an understanding of the client’s thoughts, feelings, or concerns and provides a thoughtful, insightful, or elaborate response. Complex reflections contribute to the conversation by expanding upon the client’s ideas or by offering new perspectives. 

\textbf{Empathy}: 0 (Non-Empathetic), 1 (Basic Empathy), or 2 (Advanced Empathy). Non-Empathetic (0): A response that shows no recognition or acknowledgment of the person's emotional state or perspective. E.g. Dismiss or invalidate feelings. Change the subject without addressing emotions. Offer purely factual or technical responses when emotional support is needed. Show complete misalignment with the person's emotional state Basic Empathy (1): A response that demonstrates fundamental recognition of emotions and attempts to understand the person's perspective. E.g. Acknowledge obvious or stated emotions. Use basic emotional labelling ("That must be hard"). Mirror the person's expressed feelings. Show surface-level understanding without deeper exploration. Offer general supportive statements. Advanced Empathy (2): A response that shows deep emotional attunement and sophisticated understanding of the person's experience. Connect different aspects of the person's experience and recognize nuanced emotional states. Demonstrate understanding of the broader context and implications. Show genuine emotional resonance while maintaining appropriate boundaries. Help the person gain new insights into their emotional experience.

\textbf{Fluency}: Assess the linguistic naturalness and smoothness of the counsellor’s responses. Responses are rated on a scale from 0 to 2, where 0 indicates responses that lack fluency, 1 signifies somewhat fluent responses, and 2 represents responses that are highly fluent and natural in their expression. Fluent counsellor responses should convey information in a clear and easily understandable manner, ensuring effective communication.

\end{document}